\documentclass[lettersize,journal]{IEEEtran}
\usepackage{amssymb}
\usepackage{amsthm}
\usepackage[utf8]{inputenc}
\usepackage{amsmath}
\usepackage{graphicx} 
\usepackage{multirow}
\usepackage{graphics}
\usepackage{graphicx}
\usepackage{booktabs}
\usepackage{setspace}
\usepackage{array}
\usepackage{booktabs}
\usepackage{cite}
\usepackage[colorlinks]{hyperref}
\usepackage[table,xcdraw]{xcolor}

\usepackage{makecell}
\usepackage{tikz,xcolor,hyperref}
\usepackage[caption=false,font=normalsize,labelfont=sf,textfont=sf]{subfig}
\usepackage{bm}

\usepackage{calc}
\usepackage{pgf}
\usepackage[switch]{lineno}
\usepackage{algorithm}
\usepackage{algorithmic}

\definecolor{LightRed}{rgb}{1,0.92,0.92}
\definecolor{LightOrange}{rgb}{1,0.95,0.88}
\definecolor{LightYellow}{rgb}{1.0,1.0,0.84}
\definecolor{LightGreen}{rgb}{0.9,1.0,0.88}
\definecolor{LightCyan}{rgb}{0.9,1,1}
\definecolor{LightBlue}{rgb}{0.9,0.94,1}

\definecolor{lime}{HTML}{A6CE39}
\DeclareRobustCommand{\orcidicon}{%
    \begin{tikzpicture}
    \draw[lime, fill=lime] (0,0) 
    circle [radius=0.16] 
    node[white] {{\fontfamily{qag}\selectfont \tiny ID}};    \draw[white, fill=white] (-0.0625,0.095) 
    circle [radius=0.007];    \end{tikzpicture}
    \hspace{-2mm}}
\foreach \x in {A, ..., Z}{%
    \expandafter\xdef\csname orcid\x\endcsname{\noexpand\href{https://orcid.org/\csname orcidauthor\x\endcsname}{\noexpand\orcidicon}}
    }

\begin{document}
\title{Federated Domain Generalization via Prompt Learning and Aggregation}

\author{
	Shuai Gong\orcidB{}, Chaoran Cui\textsuperscript{*}\orcidC{}, Chunyun Zhang\orcidE{}, Wenna Wang\orcidD{},  Xiushan Nie\orcidF{}, Senior Member, IEEE, and Lei Zhu\orcidG{}, Senior Member, IEEE
\thanks{This work was supported by the National Natural Science Foundation of
	China under Grant 62077033 and Grant 62072274, by the Shandong Provincial
	Natural Science Foundation under Grant ZR2020KF015, and by the Taishan
	Scholar Program of Shandong Province under Grant tsqn202211199 and Grant
	tstp20221137.
	
	S. Gong, C. Cui, C. Zhang, and Wenna Wang  are with the
	School of Computing and Artificial Intelligence, Shandong University of
	Finance and Economics, Jinan 250014, China (e-mail: gsh8210@163.com; crcui@sdufe.edu.cn;
	zhangchunyun1009@126.com; wangwenna@mail.nwpu.edu.cn.
	
	X. Nie is with the School of Computer Science and Technology,
	Shandong Jianzhu University, Jinan 250101, China (e-mail: niexsh@hotmail.com).
	
	L. Zhu is with the College of Electronics and Information Engineering,
	Tongji University, Shanghai 201804, China (e-mail: leizhu0608@gmail.com).
	}	

\thanks{\textsuperscript{*}Corresponding author.}
	}

\maketitle

\begin{abstract}
Federated domain generalization (FedDG) aims to improve the global model’s generalization in unseen domains by addressing data heterogeneity under privacy-preserving constraints.
A common strategy in existing FedDG studies involves sharing domain-specific knowledge among clients, such as spectrum information, class prototypes, and data styles. 
However, this knowledge is extracted directly from local client samples, and sharing such sensitive information poses a potential risk of data leakage, which might not fully meet the requirements of FedDG.
In this paper, we introduce prompt learning to adapt pre-trained vision-language models (VLMs) in the FedDG scenario, and leverage locally learned prompts as a more secure bridge to facilitate knowledge transfer among clients.
Specifically, we propose a novel FedDG framework through Prompt Learning and AggregatioN (PLAN), which comprises two training stages to collaboratively generate local prompts and global prompts at each federated round.
First, each client performs both text and visual prompt learning using their own data, with local prompts indirectly synchronized by regarding the global prompts as a common reference.
Second, all domain-specific local prompts are exchanged among clients and selectively aggregated into the global prompts using lightweight attention-based aggregators.
The global prompts are finally applied to adapt VLMs to unseen target domains.
As our PLAN framework requires training only a limited number of prompts and lightweight aggregators, it offers notable advantages in computational and communication efficiency for FedDG.
Extensive experiments demonstrate the superior generalization ability of PLAN across four benchmark datasets. 
We have released our code at \url{https://github.com/GongShuai8210/PLAN}.

\end{abstract}
\begin{IEEEkeywords}
Federated Learning, Federated Domain Generalization, Data Heterogeneity, Prompt Learning
\end{IEEEkeywords}

\section{Introduction}
\IEEEPARstart {I}{n} today's era of distributed data sources and edge computing, there exists a substantial demand for collaboratively training machine learning models across multiple clients.
Federated learning (FL)~\cite{mcmahan2017communication,wei2020federated} has emerged as a promising solution, enabling the development of accurate and robust models while maintaining data privacy.
Unlike traditional centralized approaches, FL allows each local client to learn from its own data.
Then, a central server periodically aggregates the local model parameters from all clients to generate a global model.

\begin{figure}[t]
	\centering
	\includegraphics[width=0.9\linewidth,height=4.33cm]{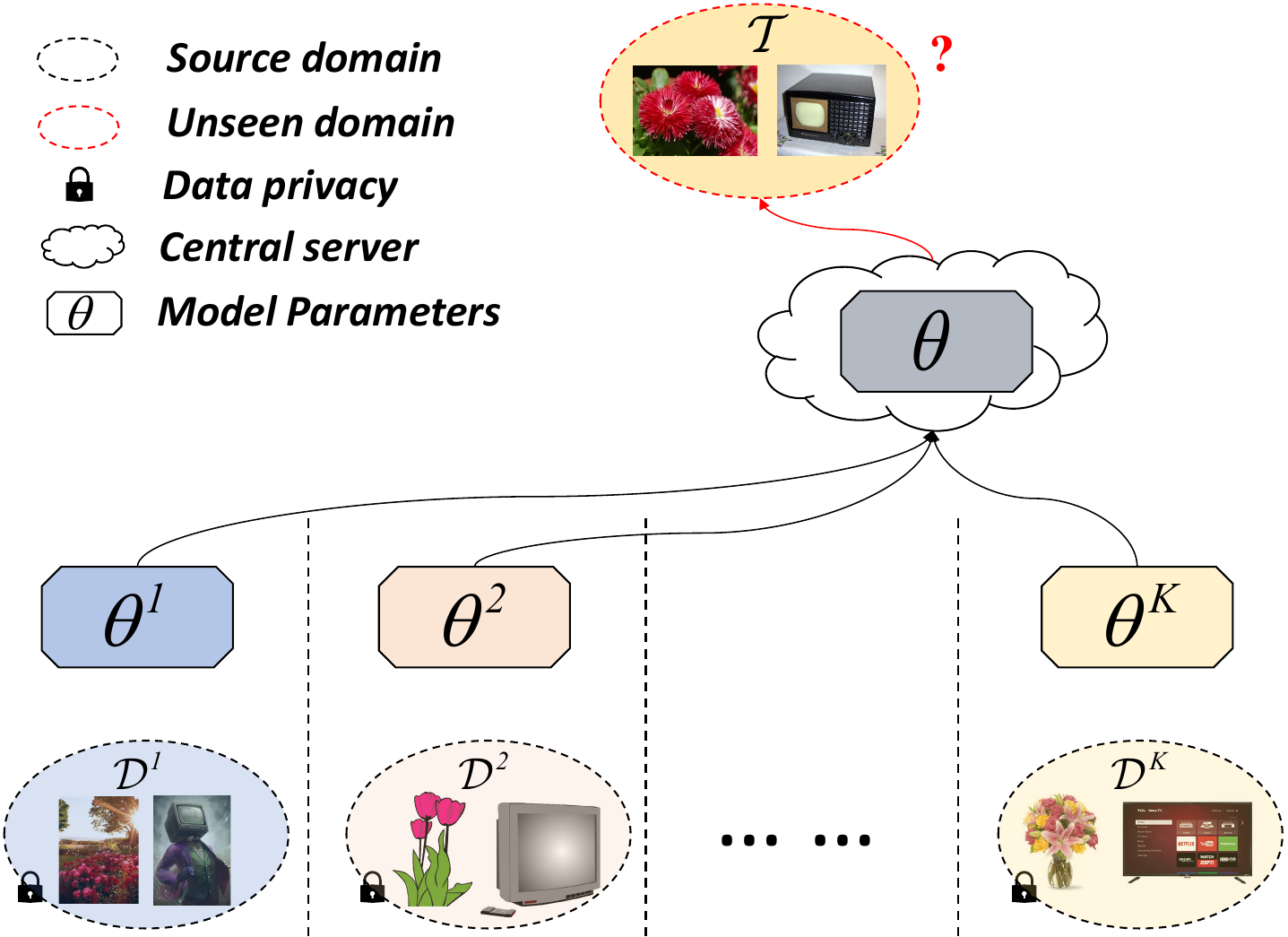}
	\caption{The novel problem setting of FedDG aims to learn a global model from multiple decentralized source domains, enabling it to directly generalize to completely unseen target domains.}
	\label{fig:head}
\end{figure}

Despite significant progress, traditional FL approaches~\cite{li2020federated,wang2020tackling,karimireddy2020scaffold,jiang2022harmofl} primarily concentrate on enhancing model performance within the federation of clients.
These methods often assume that data across all clients are identically distributed.
In reality, however, clients independently collect local data, which naturally form mutiple source domains with distinct distributions.
During deployment, the test data may also come from a target domain with previously unseen distributions. 
Therefore, how to generalize the federated model under domain shifts remains an underexplored issue.

As a common solution to the domain shift issue, domain generalization (DG)~\cite{wang2023domain,zhou2024mixstyle} techniques have been developed to enable models trained on source domains with heterogeneous distributions to generalize well to unseen target domains.
Nevertheless, applying existing DG methods in the FL setting is not straightforward, as they typically operate in a centralized manner that requires full access to data from different source domains.
To overcome this challenge, federated domain generalization (FedDG)~\cite{liu2021feddg} has been further introduced to synergize FL and DG.
As illustrated in Fig.~\ref{fig:head}, FedDG facilitates collaborative model learning across diverse source domains for effective generalization to unseen target domains while keeping data privacy.

Many prior studies on FedDG aim to share domain knowledge among clients under the privacy-preserving constraint, thereby enhancing the generalization ability of each local model by integrating knowledge from diverse domains.
For example, Liu et al.~\cite{liu2021feddg} exchanged the amplitude spectrum information in frequency space among clients.
Huang et al.~\cite{huang2023rethinking} allowed the sharing of class prototypes of local samples across different clients.
Chen et al.~\cite{chen2023federated} performed cross-client style transfer by exchanging the mean and variance of each pixel-level feature channel.
Notably, the aforementioned domain knowledge, including amplitude spectrum, class prototypes, and data style, is directly extracted from local samples and can be considered a reflection of their inherent characteristics.
Sharing such sensitive information among clients potentially poses the risk of data leakage~\cite{zhang2021federated}, which might not fully meet the requirements of FedDG.

Recently, numerous foundational vision-language models (VLMs)~\cite{radford2021learning,li2022blip} have been developed by utilizing large-scale image-text pairs from the web.
These VLMs acquire general knowledge about the associations between visual patterns and their corresponding textual descriptions.
To preserve this general knowledge and mitigate the transfer gap to downstream tasks, the technique of prompt learning~\cite{zhou2022learning,khattak2023maple} has been proposed, which freezes the parameters of VLMs while inserting a few learnable prompt tokens as additional inputs to VLMs.
In a FL scenario, prompt learning can be conducted in each client, enabling the learned prompts to encapsulate domain knowledge and facilitate the adaptation of VLMs to the tasks at hand.
It is important to note that prompts are optimized via learning rather than being generated directly from local samples in each client.
Therefore, these prompts can serve as a more secure bridge to transfer domain knowledge among clients for FedDG.


In this paper, we present a novel FedDG framework through Prompt Learning and AggregatioN (PLAN).
Our PLAN method comprises two stages in each federated round: first, learning local prompts in each client, and then, synthesizing global prompts that effectively generalize to unseen target domains.
A few recent works~\cite{guo2023promptfl, feng2023learning} have begun to explore prompt learning in the context of FL.
However, there remain two major limitations:
1) Clients independently learn local prompts, resulting in prompts biased towards each client's private data.
This bias can compromise the generalization of the global model;
and 2) A common practice for aggregating local prompts into global prompts is by using fixed weights.
However, diverse domain knowledge tends to contribute unequally to the global model, and neglecting these differences may significantly diminish model performance~\cite{yuan2021we}.

To address the above issues, PLAN introduces a reference-based prompt learning mechanism, which requires the local prompts in each client to align with a common reference---the global prompts generated in the previous round and shared among all clients.
In this way, PLAN facilitates the indirect synchronization of local prompts across clients without the need for data sharing.
Additionally, PLAN is equipped with the lightweight attention-based prompt aggregators, which measures the importance of local prompts from different clients and selectively aggregates them into the global prompts.
During the aggregation, all local prompts remain unchanged, with only the aggregators' parameters being updated.
Note that for further improved performance, PLAN performs both text and visual prompt learning and aggregation on multiple blocks in VLMs.
In each federated round, clients only need to train and exchange a limited set of prompt tokens and the lightweight aggregators with the server, rather than the entire model.
Therefore, PLAN not only offers robust privacy-preserving capability but also enhances computational and communication efficiency for FedDG.

Experimental results on four benchmark datasets demonstrate that our PLAN method significantly outperforms conventional FL, DG, and FedDG methods, as well as prompt learning-based methods.
Ablation studies and hyperparameter analyses validate the key components of PLAN.
We further verify the effectiveness of PLAN when faced with insufficient local data and confirm its advantages in both computational and communication costs.
Visualization studies are also conducted to provide deeper insights into PLAN.

In a nutshell, our main contributions are as follows:

\begin{itemize}
	
	\item Instead of sharing domain knowledge directly extracted from local samples, we introduce prompts learned with local data as a more secure means of transferring domain knowledge among clients in FedDG. 
	
	\item We propose a novel FedDG framework, PLAN, which integrates a reference-based prompt learning mechanism to facilitate the indirect synchronization of local prompts across clients and utilizes the lightweight attention-based prompt aggregators to selectively aggregate local prompts into global prompts. 
	
	\item We conduct extensive experiments on four benchmark datasets and demonstrate that our PLAN method achieves new state-of-the-art performance while maintaining communication and computation efficiency for FedDG.
\end{itemize}

The remainder of the paper is structured as follows:
Section~\ref{sec:related_work} provides a review of related work.
Section~\ref{sec:preliminary} presents the preliminary knowledge relevant to our study.
Section~\ref{sec:method} details our PLAN method for FedDG.
Experimental results and analysis are discussed in Section~\ref{sec:experiments}, followed by the conclusions in Section~\ref{sec:conclusions}.

\section{Related Work}\label{sec:related_work}

\subsection{Federated Learning}
Federated learning (FL) is a distributed machine learning paradigm that enables multiple clients to collaboratively train a model while keeping their data localized. 
In a single iteration of FL, the server sends the current global model to all participating clients. Subsequently, each client independently trains the model using its local dataset. Once the local training is completed, each client uploads its model parameters back to the server, allowing for the aggregation of these parameters to form a new global model.

FedAvg~\cite{mcmahan2017communication} is a fundamental FL algorithm that aggregates locally trained model parameters through weighted averaging, with weights determined by each client’s data size. 
However, due to its underlying assumption of the IID data distribution, FedAvg often encounters difficulties in settings characterized by data heterogeneity among clients. 
To mitigate the challenges posed by data heterogeneity in FL,
FedProx~\cite{li2020federated} extends FedAvg by incorporating a proximal term into the local objective functions, which prevents local updates from deviating too much from the global model.
MOON~\cite{li2021model} align the feature representations learned by the local model and global model through the  contrastive learning objective. 
Instead of aligning local updates with the global model,
SCAFFOLD~\cite{karimireddy2020scaffold} mitigates client drift by employing control variates to adjust and correct local updates.
FedNova~\cite{wang2020tackling} tackles objective inconsistency resulting from heterogeneous data and varying local computations by normalizing local updates before aggregation.
FedBN~\cite{li2021fedbn} addresses feature distribution heterogeneity by keeping local batch normalization layers, allowing clients to maintain personalized feature statistics.
In addition to these local optimization methods, FedOpt~\cite{reddi2020adaptive} applies adaptive optimization methods on the server side to improve convergence.
Although these FL studies effectively address distributional disparities among local clients, they overlook a common scenario in which the data distribution of the target dataset significantly diverges from that of the local clients.

\subsection{Domain Generalization}

The above issue can be addressed by the domain generalization (DG) technique. 
DG aims to train a model on one or more source domains that can generalize well to the unseen target domain.
Representative methods either learn domain-invariant features across multiple source domains\cite{zhou2024mixstyle,dayal2024madg,lin2024diversifying,liu2023robust} or adopt the idea of meta-learning\cite{li2018learning,qin2023bi}.
Most conventional DG methods train  models in a centralized manner. However, these methods cannot be directly applied in FL, since the centralized server is restricted from accessing multi-source domains to comply with data privacy requirements.

Such limitation of centralized methods  leads to the emergence of FedDG. Federated domain alignment is crucial in FedDG, as it aims to reduce domain shift and enhance the model's generalization ability on unseen domains \cite{li2023federated}. 
To achieve this, a core idea is to facilitate domain knowledge sharing across multiple clients.
Liu et al.~\cite{liu2021feddg} exchanged the amplitude spectrum information in frequency space among clients to transfer multi-client data distribution. Huang et al.~\cite{huang2023rethinking}  allows the sharing of  class prototypes of local data among clients. Chen et al.~\cite{chen2023federated} introduced CCST, which facilitates image style sharing by exchanging pixel-level channel-wise mean and standard deviation of image features among clients.
The shared information of these methods is generated directly from the local data.
Consequently, sharing such sensitive information among clients could be considered as a form of data leakage. Our approach, PLAN, introduces prompts learned from local data as a more secure bridge for transferring domain knowledge among clients in FedDG. 

\subsection{Prompt Learning in VLMs}
Prompt learning~\cite{zhou2022conditional,khattak2023self} is a kind of parameter-efficient fine-tuning techniques, which only introduces a small number of learnable prompt tokens as extra inputs while freezing the parameters of VLMs. 
As a pioneering effort, CoOp~\cite{zhou2022learning} converts the input context tokens in the text encoder into learnable prompts for adapting CLIP~\cite{radford2021learning} to downstream image recognition. MaPLe~\cite{khattak2023maple} further introduce visual prompts to enhance alignment between the text and visual representations in CLIP.
In recent years, researchers have introduced  the prompt learning methods into FL because they naturally meet the requirements for efficient communication.
For example, PromptFL~\cite{guo2023promptfl} and Diprompt~\cite{bai2024diprompt} independently learn text prompts on each local client, then aggregate the locally trained prompts at the server with fixed weights. In contrast, PLAN concurrently performs both text and visual prompt learning on the local client through a reference-based prompt learning mechanism, which facilitates the indirect synchronization of local prompts across clients. Additionally, PLAN is equipped with  lightweight, attention-based prompt aggregators to selectively aggregate local prompts into global prompts.

\section{Preliminaries}\label{sec:preliminary}
In this section, we first present the formulation of the FedDG problem. 
We then briefly describe CLIP~\cite{radford2021learning}, a typical representative of VLMs.

\subsection{Federated Domain Generalization}
FedDG aims to develop models that can generalize well to unseen domains while preserving data privacy across distributed sources. 
As a common practice in FedDG, we generally consider the $C$-way image classification task, where heterogeneous data is partitioned among $K$ clients. 

Formally, we denote $(\mathcal{X}, \mathcal{Y})$ as the joint image and label space.
Each client independently collects local data, resulting in data heterogeneity among clients and thus forming multiple source domains.
Let $\mathcal{S}=\{\mathcal{S}^k\}_{k=1}^{K}$ represent the dataset of $K$ clients, where $\mathcal{S}^k$ consists of ${N_k}$ image and label pairs for client $k$, i.e., $\mathcal{S}^k=\{(\bm{x}^{k}_i,y^{k}_i) \in (\mathcal{X}, \mathcal{Y})\}_{i=1}^{N_k}$.
For any two distinct clients $k$ and $j$, their data distributions are different: 
$
\mathbb{P}_{\mathcal{S}^k} \neq \mathbb{P}_{\mathcal{S}^{j}}, \forall k, j \in \{1, \dots, K\}.
$
FedDG aims to learn a global model $f:\mathcal{X} \rightarrow \mathcal{Y}$ from the combined dataset $\mathcal{S}$, with the goal of achieving high performance in unseen target domains.

The standard FedDG paradigm involves two key stages: (1) \textit{local training}, where each client trains a model on its own domain-specific data to capture local patterns, and (2) \textit{global aggregation}, where the locally trained models are sent to a central server for aggregation, producing a global model that generalizes to unseen domains. 
Typically, these two stages alternate multiple times, with each full cycle referred to as \emph{a federated round}.

\subsection{Revisiting CLIP}
Our work focuses on effectively adapting VLMs through prompt learning in the FedDG scenario.
We use CLIP~\cite{radford2021learning} as the VLM of choice due to its broad popularity.
CLIP comprises a text encoder and an image encoder. 
We utilize a Transformer~\cite{vaswani2017attention} as the text encoder. 
Since visual prompt learning is also considered, a Vision Transformer (ViT)~\cite{dosovitskiy2020image} is selected as the image encoder.

The text encoder processes a category description, typically in the format ``a photo of a [CLASS].'', where [CLASS] denotes the category name. 
The description is first tokenized into a sequence of tokens and projected into word embeddings ${\bm{E}_0}$.
The text encoder then generates the text representation $\bm{w}$ for the given category.
The vision encoder takes an image as input, which is split into fixed-size patches and mapped into patch embeddings ${\bm{P}_0}$. 
The vision encoder generates the visual feature $\bm{f}$ for the image.

CLIP is trained to match images with their corresponding categories using a contrastive loss~\cite{liu2021simcls}.
Specifically, the text encoder generates the text representation $\bm{w}_c$ for each category $c \in \left\{1, 2, \ldots, C\right\}$.
Given an image $\bm{x}$ with its visual feature $\bm{f}$, CLIP predicts the probability that $\bm{x}$ belongs to category $c$ as follows:
\begin{equation}\label{eq:clip}
	p\left( {y = c|\bm{x}} \right) = \frac{{\exp \left( {\cos \left( {{\bm{w}_c},\bm{f}} \right)/\tau } \right)}}{{\sum\nolimits_{j = 1}^C {\exp \left( {\cos \left( {{\bm{w}_j},\bm{f}} \right)/\tau } \right)} }},
\end{equation}
where $\tau$ is a learnable temperature parameter, and $\cos(\cdot, \cdot)$ denotes the cosine similarity.
\section{Method}\label{sec:method}
\begin{figure*}[t]
	\centering
	\includegraphics[width=0.95\linewidth]{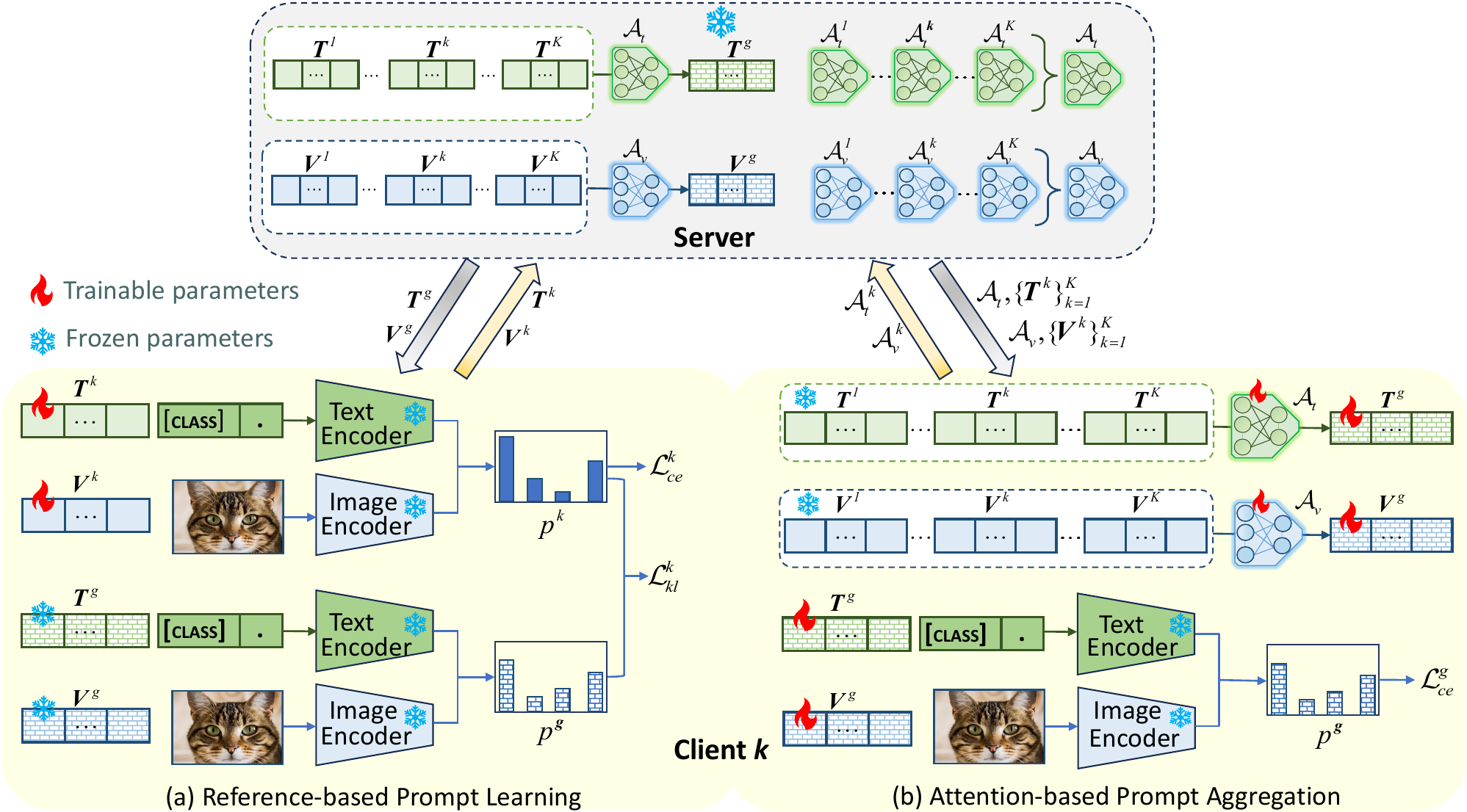}
	\caption{Overall framework of PLAN for FedDG.  PLAN consists of two training stages at each federated round. In stage (a), the server sends the global text and visual prompts, denoted as $\bm{T}^g$ and $\bm{V}^g$, to each client. Client $k$ then learns its own text and visual prompts, $\bm{T}^k$ and $\bm{V}^k$, using local data and aligns these learned prompts with the global prompts. The learned prompts are subsequently uploaded to the server. In stage (b), the server distributes the global aggregators, $\mathcal{A}_t$ and $\mathcal{A}_v$, along with all clients' local prompts to each client. Each client then learns to selectively aggregate local prompts from all clients into the global prompts, and the learned aggregators, $\mathcal{A}_t^k$ and $\mathcal{A}_v^k$, are uploaded back to the server.} 
	\label{fig:PLAN}
\end{figure*}
In this section, we elaborate our proposed FedDG framework based on Prompt Learning and AggregatioN (PLAN).
PLAN simultaneously performs both text and visual prompt learning for CLIP on all clients and leverages the learned prompts as a secure bridge for transferring domain knowledge among clients.
Following the standard FedDG paradigm, PLAN integrates two training stages in each federated round: first, a reference-based prompt learning mechanism is introduced to enable the indirect synchronization of local prompts across clients; second, the lightweight attention-based prompt aggregators are implemented to selectively aggregate local prompts into global prompts.
The global prompts are finally used to adapt CLIP to unseen target domains.
For clarity, Table~\ref{tbl:notation} presents a summary of the key notations and their definitions used throughout this paper. 
The conceptual structure of PLAN is illustrated in Fig.~\ref{fig:PLAN}. 

\begin{table}[t]
	\centering
	\renewcommand{\arraystretch}{1.25} 
	
	\caption{Overview of Key Notations and Correspon Definitions.}\label{tbl:notation}
	\footnotesize 
	\begin{tabular}{>{\raggedright\arraybackslash}p{0.33\columnwidth} >{\raggedright\arraybackslash}p{0.6\columnwidth}}
		\toprule
		\textbf{Notation} & \textbf{Definition} \\
		\midrule
		$\mathcal{S}=\{\mathcal{S}^k\}_{k=1}^{K}$& The dataset of $K$ clients \\
		$\mathcal{S}^k=\{\bm{x}^{k}_i,y^{k}_i\}_{i=1}^{N_k}$ &Labeled samples in the client $k$ \\
		$\left\{ \mathcal{B}_l \right\}_{l = 0}^{L-1}$ & Text encoder of CLIP with $L$ Transformer blocks \\
		$\{ \mathcal{I}_l \}_{l=0}^{L-1}$ & Vision encoder of CLIP with $L$ Vision Transformer blocks \\
		$\bm{E}^{c}_{l}$ & Word embeddings for the description of category $c$ fed into $\mathcal{B}_l$ \\
		$\bm{P}_l$ & Image patch embeddings fed into $\mathcal{I}_l$\\
		$\bm{T}^k, \bm{T}^g$ &  Text prompts in the client $k$ and global text prompts\\
		$\bm{w}_c^k, \bm{w}_c^g$ &  Text representation of category $c$ generated with $\bm{T}^k$ and  $\bm{T}^g$\\
		$\bm{V}^k, \bm{V}^g$ &  Visual prompts in the client $k$ and global visual prompts\\
		$\bm{f}_i^k,\bm{f}_i^g$  &  Visual features of image $\bm{x}_i^k$  extracted with the visual prompts $\bm{V}^k$ and  $\bm{V}^g$\\
		$p_c^k( \bm{x}_i^k )$ & The probability of $\bm{x}_i^k$ being classified into category $c$ derived from $\bm{T}^k$ and  $\bm{V}^k$  \\
		$ p_c^g( \bm{x}_i^k)$ & The probability of $\bm{x}_i^k$ being classified into category $c$ derived from $\bm{T}^g$ and  $\bm{V}^g$\\
		$\mathcal{A}_t^k$,$\mathcal{A}_t$ &Local text aggregator in the client $k$ and the global text aggregator\\
		$\mathcal{A}_v^k$,$\mathcal{A}_v$ &Local visual aggregator in the client $k$ and the global visual aggregator \\
		\bottomrule
	\end{tabular}
\end{table}
\subsection{Prompt Designing}
Prior research~\cite{khattak2023maple} has confirmed the advantages of multi-modal prompts in guiding VLMs.
Additionally, appending prompts to the deeper encoder blocks helps to the learning of stage-wise feature representations.
Building on these insights, we incorporate both text and visual prompts into each block of the CLIP encoders.

Suppose the text encoder comprises $L$ Transformer blocks $\left\{ \mathcal{B}_l \right\}_{l = 0}^{L-1}$.
For client $k$, we design $m_t$ text prompt tokens for block $\mathcal{B}_l$, denoted as $\bm{T}_l^k = {\left[ {\bm{t}_l^k} \right]_1}{\left[ {\bm{t}_l^k} \right]_2} \ldots {\left[ {\bm{t}_l^k} \right]_{{m_t}}}$.
These tokens are inserted into $\mathcal{B}_l$ on client $k$ and computed as follows:
\begin{equation}\label{eq:text_encoder}
	{{\left[ \texttt{cls} \right]_{l+1}}, \underline{\hspace{1em}}, {\bm{E}^{c}_{l+1}}} = {\mathcal{B}_l}\left( { {{\left[ \texttt{cls} \right]_l},{\bm{T}_l^k}, {\bm{E}^{c}_{l}}} } \right),
\end{equation}
where ${\left[ \texttt{cls} \right]_l}$ denotes the input class token for $\mathcal{B}_l$, and ${\bm{E}^{c}_{l}}$ denotes the embedding of the description of category $c$.
Finally, the text representation of category $c$ on client $k$ is generated by
\begin{equation}\label{eq:s_weights}
	\bm{w}_c^k = {\mathop{\rm Proj}\nolimits_t} \left( {{\left[ \texttt{cls} \right]_L}} \right),
\end{equation}
where ${\mathop{\rm Proj}\nolimits_t}\left( \cdot \right)$ is a projection layer for the text encoder.

The vision encoder also consists of $L$ Transformer blocks $\left\{ \mathcal{I}_l \right\}_{l = 0}^{L-1}$.
For client $k$, we insert $m_v$ visual prompt tokens $\bm{V}_l^k = {\left[ {\bm{v}_l^k} \right]_1}{\left[ {\bm{v}_l^k} \right]_2} \ldots {\left[ {\bm{v}_l^k} \right]_{{m_v}}}$ into block $\mathcal{I}_l$.
The visual prompt tokens are processed by
\begin{equation}\label{eq:visual_encoder}
	{{\left[ \texttt{cls} \right]_{l+1}}, \underline{\hspace{1em}}, {\bm{P}_{l+1}}} = {\mathcal{I}_l}\left( { {{\left[ \texttt{cls} \right]_l},{\bm{V}_l^k}, {\bm{P}_{l}}} } \right),
\end{equation}
where ${\left[ \texttt{cls} \right]_l}$ and ${\bm{P}_{l}}$ are the class token and the image patch embeddings fed into $\mathcal{I}_l$, respectively.
A linear projection layer then maps ${\left[ \texttt{cls} \right]_L}$ into the final visual feature of the input image for client $k$:
\begin{equation}\label{eq:visual_features}
	\bm{f}^k = {\mathop{\rm Proj}\nolimits_v} \left( {{\left[ \texttt{cls} \right]_L}} \right).
\end{equation}

Note that unlike existing studies~\cite{bai2024diprompt,zhou2022conditional}, our PLAN method randomly initializes prompt tokens on each client without relying on any domain-specific information.
This is necessary because the learned prompts will be shared across clients for FedDG, and directly injecting domain-specific information could increase the risk of privacy leakage for the client~\cite{wen2024hard}.
PLAN simultaneously learns the text and visual prompts $\bm{T}_l^k$ and $\bm{V}_l^k$ for each block $\mathcal{B}_l$ and $\mathcal{I}_l$ on client $k$.
In the following sections, we will focus on describing the learning and aggregation processes for a specific block. 
For simplicity, the subscript $l$ will be omitted when no ambiguity arises.

\subsection{Reference-based Prompt Learning}~\label{sec:reference-based}
For local training, our PLAN method updates the local text and visual prompts using domain-specific data on each client.
Specifically, given a labeled sample from client $k$, i.e., $(\bm{x}^{k}_i,y^{k}_i) \in \mathcal{S}^k$, the probability distribution of image $\bm{x}^{k}_i$ over categories, denoted as $\bm{p}^k(\bm{x}_i^k) \in [0, 1]^C$, where the $c$-th element represents the probability of $\bm{x}_i^k$ being classified into category $c$, is predicted as follows:
\begin{equation}\label{eq:p_s}
	p_c^k( \bm{x}_i^k ) = \frac{{\exp \left( {\cos ( {{\bm{w}_c^k},\bm{f}_i^k} )/\tau } \right)}}{{\sum\nolimits_{j = 1}^C {\exp \left( {\cos ({{\bm{w}_j^k},\bm{f}_i^k} )/\tau } \right)} }}.
\end{equation}
Here, ${\bm{w}_c^k}$ denotes the text representation of category $c$ generated with the text prompts $\bm{T}^k$ (refer to Eq.~\eqref{eq:text_encoder} and Eq.~\eqref{eq:s_weights}), while $\bm{f}_i^k$ denotes the features of image $\bm{x}_i^k$ extracted with the visual prompts $\bm{V}^k$ (refer to Eq.~\eqref{eq:visual_encoder} and Eq.~\eqref{eq:visual_features}). 
Then, $\bm{T}^k$ and $\bm{V}^k$ can be optimized by minimizing the cross-entropy loss: 
\begin{equation}\label{eq: s_ce}
	\mathcal{L}_{ce}^k =  - \frac{1}{{{N_k}}}\sum\limits_{i = 1}^{{N_k}} {\log \left( {p_{y_i^k}^k( {\bm{x}_i^k} )} \right)}.
\end{equation}

Current prompt learning-based FedDG methods~\cite{bai2024diprompt,su2024federated} primarily rely on the cross-entropy loss to learn local prompts. 
However, this approach leads to each client conducting local training entirely independently, resulting in prompts that are biased towards the client's private data. 
Such bias has been demonstrated to diminish the inherent generalization ability of the CLIP model when applied to new domains~\cite{khattak2023self}.
To address this issue, PLAN introduces a reference-based prompt learning mechanism, which seeks to align the local prompts of different clients with a common reference.
In this way, local prompts can be indirectly synchronized without requiring data sharing, thereby mitigating the potential domain shift caused by data isolation in distributed clients.

In particular, we selected the global prompts generated in the previous federated round as the common reference.
This choice is motivated by the fact that the global prompts are aggregated from various local prompts, thus preserving cross-client domain knowledge.
The stage of generating the global prompts will be detailed in Section~\ref{sec:prompt_aggregation}.
Let $\bm{T}^g$ and $\bm{V}^g$ denote the global text and visual prompts, respectively, which are distributed by the server and inserted into blocks $\mathcal{B}$ and $\mathcal{I}$ on client $k$.
Similar to ${\bm{w}_c^k}$ and $\bm{f}_i^k$, we can derive the global text representation ${\bm{w}_c^g}$ with $\bm{T}^g$ for category $c$, as well as the global visual feature $\bm{f}_i^g$ using $\bm{V}^g$ for image $\bm{x}_i^k$.
Based on the global prompts, the probability that $\bm{x}_i^k$ belongs to category $c$ is computed by
\begin{equation}\label{eq:p_g}
	p_c^{g}\left( \bm{x}_i^k \right) = \frac{{\exp \left( {\cos ( {{\bm{w}_c^g},\bm{f}_i^{g}} )/\tau } \right)}}{{\sum\nolimits_{j = 1}^C {\exp \left( {\cos( {{\bm{w}_j^g},\bm{f}_i^{g}})/\tau } \right)} }}.
\end{equation}
To align the local prompts on client $k$ with the global prompts, we minimize the Kullback-Leibler (KL) divergence~\cite{kullback1951information} between their corresponding prediction distributions:
\begin{equation}\label{eq:kl_loss}
	{\cal L}_{kl}^k = \frac{1}{{{N_k}}}\sum\limits_{i = 1}^{{N_k}} {\sum\limits_{j = 1}^C {p_j^g\left( {\bm{x}_i^k} \right)\log \left( {\frac{{p_j^g\left( {\bm{x}_i^k} \right)}}{{p_j^k\left( {\bm{x}_i^k} \right)}}} \right)} }.
\end{equation}
During local training, this loss encourages all local prompts to mimic the shared global prompts, thereby facilitating their indirect synchronization.
Moreover, it can be interpreted as a form of label smoothing regularization~\cite{yuan2020revisiting}, where the global prompts are used to regularize the training of local prompts, thus enhancing their generalization.

Note that since the global prompts are generated by aggregating diverse local prompts, they only become available after the first federated round ends.
Therefore, during the initial round, we leverage the zero-shot inference capability of CLIP and take its prediction distribution as a substitute for calculating the KL loss as shown in Eq.~\eqref{eq:kl_loss}.  
To sum up, we optimize the local prompts  $\bm{T}^k$ and $\bm{V}^k$ on each client $k$ by combining the cross-entropy loss and the KL loss:
\begin{equation}\label{eq: t_ce}
	\mathcal{L}_{}^k =   \mathcal{L}_{ce}^k + \alpha \mathcal{L}_{kl}^k,
\end{equation}
where $\alpha$ is a hyperparameter balancing the two losses.
In this optimization, the global prompts $\bm{T}^g$ and $\bm{V}^g$ are kept fixed.
Once local training has been completed, each client uploads its local prompts to the server, where they are prepared for aggregation into the new global prompts.

\subsection{Attention-based Prompt Aggregation}~\label{sec:prompt_aggregation}
For global aggregation, a typical approach is to average the local prompts to obtain the global prompts~\cite{guo2023promptfl,feng2023learning}.
However, in the context of FedDG, local prompts encapsulate diverse domain knowledge from various clients.
Simply assuming equal contribution from each local prompt to the global prompts may compromise the model's generalization to unseen domains.
To overcome this challenge, our PLAN method introduces  two lightweight attention-based aggregators to measure the importance of local prompts and selectively aggregate them into the global prompts.

From all clients, the server collects the set of local text prompts $\{\bm{T}^k\}_{k=1}^{K}$ and local visual prompts $\{\bm{V}^k\}_{k=1}^{K}$.
Our objective is to learn two corresponding prompt aggregators, $\mathcal{A}_t$ and $\mathcal{A}_v$, which effectively aggregate the local prompts into the global text prompt $\bm{T}^g$ and global visual prompt $\bm{V}^g$, respectively:
\begin{equation}\label{eq:aggregator}
	\begin{aligned}
		\bm{T}^g &= \mathcal{A}_t \left(\bm{T}^1, \bm{T}^2, \ldots, \bm{T}^K \right), \\
		\bm{V}^g &= \mathcal{A}_v \left(\bm{V}^1, \bm{V}^2, \ldots, \bm{V}^K \right).
	\end{aligned}
\end{equation}
Since $\mathcal{A}_t$ and $\mathcal{A}_v$ follow similar procedures, we will use $\mathcal{A}_t$ as an example to outline the process of prompt aggregation.
Specifically, to capture the differences in the impacts of $\{\bm{T}^k\}_{k=1}^{K}$, $\mathcal{A}_t$ quantifies the attention weight allocated to each local text prompt $\bm{T}^k$ by assessing its degree of similarity to a learnable query $\bm{Q}$:


\begin{equation}
	\gamma_k = \frac{\exp\left(\left\langle {\bm{Q}, \mathcal{F}_q\left(\bm{T}^k\right)}\right\rangle\right)}{\sum_{j=1}^{K} \exp\left(\left\langle {\bm{Q}, \mathcal{F}_q\left(\bm{T}^j\right)}\right\rangle\right)},
	\label{eq:atten}
\end{equation}
where $\left\langle {\cdot, \cdot} \right\rangle$ denotes the inner product similarity, and $\mathcal{F}_q$ serves as a transformation function that maps $\bm{T}^k$ into the same space as $\bm{Q}$.
For computational efficiency, $\mathcal{F}_q$ is designed as a simple two-layer perceptron with a bottleneck structure~\cite{he2016deep}.
The global text prompt $\bm{T}^g$ is then generated via the weighted aggregation of the local prompts:
\begin{equation}
	\bm{T}^g=\sum\limits_{k = 1}^{K} \gamma_k \mathcal{F}_a\left(\bm{T}^k\right),
	\label{eq:aggre}
\end{equation}
where $\mathcal{F}_a$ is a mapping function that shares the same structure as $\mathcal{F}_q$.
Through this mechanism, the text aggregator $\mathcal{A}_t$ selectively emphasizes informative local prompts while suppressing less relevant ones in forming $\bm{T}^g$.
Note that $\mathcal{A}_t$ has a lightweight architecture consisting of only a small number of learnable parameters, including $\bm{Q}$, $\mathcal{F}_q$, and $\mathcal{F}_a$.
In a similar manner to $\mathcal{A}_t$, the visual aggregator $\mathcal{A}_v$ can assemble the global visual prompt $\bm{V}^g$ from the set  $\{\bm{V}^k\}_{k=1}^{K}$.


To guarantee the model guided by the aggregated global prompts generalizes effectively to unseen domains, it is crucial to optimize the prompt aggregation process for clients with diverse domain distributions.
Therefore, during the global aggregation stage, the server distributes to each client both the sets of local prompts $\{\bm{T}^k\}_{k=1}^{K}$ and $\{\bm{V}^k\}_{k=1}^{K}$, as well as the parameters of the aggregators $\mathcal{A}_t$ and $\mathcal{A}_v$.
As previously mentioned, exchanging local prompts between clients does not introduce the risk of data leakage.
On each client $k$, the parameters of $\mathcal{A}_t$ and $\mathcal{A}_v$ are updated based on the local dataset $\mathcal{S}^k$.
Specifically, client $k$ generates the global prompts $\bm{T}^g$ and $\bm{V}^g$ using $\mathcal{A}_t$ and $\mathcal{A}_v$, respectively.
$\bm{T}^g$ is applied to derive the global text representation ${\bm{w}_c^g}$ for category $c$, while $\bm{V}^g$ serves to construct the global visual feature $\bm{f}_i^g$ for image $\bm{x}_i^k \in\mathcal{S}^k$.
The probability $p_c^{g}$ of $\bm{x}_i^k$ being classified into category $c$ can be calculated as shown in Eq.~\eqref{eq:p_g}.
By training on the classification task with the cross-entropy loss, client $k$ locally optimizes $\mathcal{A}_t$ and $\mathcal{A}_v$ with its own data:
\begin{gather}\label{eq:g_ce}
	\mathcal{L}_{ce}^g =  - \frac{1}{{{N_k}}}\sum\limits_{i = 1}^{{N_k}} {\log \left( {p_{y_i^k}^g( {\bm{x}_i^k} )} \right)}, \\
	\label{eq:ag}
	{{\mathcal{A}^k_{t}}, {\mathcal{A}^k_{v}}} = \mathop {\arg\min }\limits_{\mathcal{A}_t, \mathcal{A}_v} \mathcal{L}_{ce}^g.
\end{gather}
During the optimization of the aggregators $\mathcal{A}_t$ and $\mathcal{A}_v$, all local prompts distributed to client $k$ remain unchanged.

Finally, all $K$ clients upload their local aggregator parameters to the server, where they are averaged to form the new version of text and visual aggregators:
\begin{equation}\label{eq:average}
	\mathcal{A}_t = \frac{1}{K}\sum\limits_{k = 1}^K {\mathcal{A}^k_t},\ \mathcal{A}_v = \frac{1}{K}\sum\limits_{k = 1}^K {\mathcal{A}^k_v}.
\end{equation}
As described in Eq.~\eqref{eq:atten} and Eq.~\eqref{eq:aggre}, the server once again invokes $\mathcal{A}_t$ and $\mathcal{A}_v$, producing the ultimate global text prompt $\bm{T}^g$ and visual prompt $\bm{V}^g$ for this round, respectively.
Note that although the aggregators $\mathcal{A}_t$ and $\mathcal{A}_v$ are obtained through a simple averaging of their local counterparts across clients, they are capable of performing a weighted aggregation of local prompts into the global prompts and enabling the resulting model to achieve superior cross-domain generalization.

Overall, our PLAN method involves two communications between the server and clients within each federated round:
During local training, the server distributes fixed global prompts to each client as the common reference for synchronizing local prompts, and each client subsequently returns its learned local prompts to the server.
During global aggregation, the server sends the full set of local prompts, along with the prompt aggregators to each client, enabling them to optimize the aggregators to better adapt the global prompts to their local data.
All clients then upload their locally optimized aggregators, which are combined on the server into the global aggregators to generate the final global prompts.
The training process is presented in Algorithm~\ref{alg:training}.

\begin{algorithm}[t]
	\caption{The training process of PLAN.}
	\label{alg:training}
	\begin{algorithmic}[0]
		\renewcommand{\algorithmicrequire}{ \textbf{Input:}} 
		\REQUIRE ~~\\
		\renewcommand{\arraystretch}{1}
		Datasets of $K$ clients $\{\mathcal{S}^k\}_{k=1}^{K}$, 
		pre-trained CLIP model, number of federated rounds $R$, and number of local update epochs $E$.
		\renewcommand{\algorithmicrequire}{ \textbf{Output:}}
		\REQUIRE ~~\\
        Global text prompts $\bm{T}^g$ and global visual prompts $\bm{V}^g$. 
		\FOR{$r = 1,2,..., R$}
		\STATE
		\hspace{-1em}\textcolor[rgb]{0.5,0.5,0.5}{/* Reference-based Prompt Learning */}
		\STATE
		\textbf{Server:}
		\STATE \hspace{1em} Send $\bm{T}^g$ and $\bm{V}^g$ to clients.
		\STATE \textbf{Client:}
		\STATE \hspace{1em} Initialize: $\bm{T}^k=\bm{T}^g$ and $\bm{V}^k=\bm{V}^g$.
		\STATE \hspace{1em} \textbf{for} {$e = 1,2,..., E$} \textbf{do}
		\STATE \hspace{2em} Update $\bm{T}^k$ and $\bm{V}^k$ based on $\mathcal{L}^k$ . \textcolor[rgb]{0.5,0.5,0.5}{// Eq.~\eqref{eq: t_ce}}
        \STATE \hspace{1em} \textbf{end for}
		\STATE \hspace{1em}  Upload $\bm{T}^k$ and $\bm{V}^k$ to the server.
		\STATE
		\hspace{-1em}\textcolor[rgb]{0.5,0.5,0.5}{/* Attention-based Prompt Aggregation */}
		\STATE 
		\textbf{Server:}
		\STATE 
		\hspace{1em} Send $\{\bm{T}^k\}_{k=1}^{K}$, $\{\bm{V}^k\}_{k=1}^{K}$, $\mathcal{A}_t$, and $\mathcal{A}_v$ to clients.
		\STATE \textbf{Client:}
		\STATE 
		\hspace{1em} Aggregate: $\bm{T}^g = \mathcal{A}_t \left(\bm{T}^1, \bm{T}^2, \ldots, \bm{T}^K \right)$, 
        \STATE \hspace{5.6em} $\bm{V}^g = \mathcal{A}_v \left(\bm{V}^1, \bm{V}^2, \ldots, \bm{V}^K \right)$.
		\STATE \hspace{1em} \textbf{for} {$e = 1,2,..., E$} \textbf{do}
		\STATE
		\hspace{2em} Update $\mathcal{A}_t^k $ and $\mathcal{A}_v^k$ based on $\mathcal{L}_{ce}^g$. \textcolor[rgb]{0.5,0.5,0.5}{// Eq.~\eqref{eq:ag}}
		\STATE \hspace{1em} \textbf{end for}
		\STATE
		\hspace{1em} Upload $\mathcal{A}_t^k$ and $\mathcal{A}_v^k$ to the server.
		\STATE
		\textbf{Server:}
		\STATE
        \hspace{1em} Aggregate: $\mathcal{A}_t = \frac{1}{K}\sum\nolimits_{k = 1}^K {\mathcal{A}^k_t},\ \mathcal{A}_v = \frac{1}{K}\sum\nolimits_{k = 1}^K {\mathcal{A}^k_v}$.
		\STATE 
		\hspace{1em} Aggregate: $\bm{T}^g = \mathcal{A}_t \left(\bm{T}^1, \bm{T}^2, \ldots, \bm{T}^K \right)$, 
        \STATE \hspace{5.6em} $\bm{V}^g = \mathcal{A}_v \left(\bm{V}^1, \bm{V}^2, \ldots, \bm{V}^K \right)$.
		\ENDFOR
	\end{algorithmic}
\end{algorithm}

\section{Experiments}\label{sec:experiments}
In this section, we offer a detailed description of the experimental settings and report a series of experimental results, demonstrating the effectiveness of our PLAN method.

\subsection{Datasets} 
We evaluate the performance of PLAN on four DG benchmark datasets, including:
\begin{itemize}

	\item \textbf{PACS}~\cite{li2017deeper} contains 9,991 images  across 7 classes, spanning four domains: Photo, Sketch, Cartoon, and Art Paintings.
	
	\item \textbf{OfficeHome}~\cite{venkateswara2017deep}  consists of 24,788 images in 65 classes, distributed among four domains: Art, Clipart, Product, and Real World. 
	
	\item \textbf{VLCS}~\cite{fang2013unbiased} integrates four photographic domains, i.e., VOC2007, LabelMe, Caltech-101, and SUN09, and comprises 10,729 instances across five classes.

	\item \textbf{DomainNet}~\cite{peng2019moment} combines six photographic domains---Clipart, Infograph, Painting, Quickdraw, Real World, Sketch---and includes 569,010 images in 345 classes.  
\end{itemize}

We conducted evaluations on all datasets using the leave-one-domain-out setting \cite{zhang2023federated}, where each domain is chosen as the unseen target domain for testing, while the remaining domains are allocated to different clients as source domains for training.
The split of training and validation sets within each source domain follows the same configuration as described in~\cite{gulrajanisearch,huang2020self,xu2021fourier} for PACS, OfficeHome, and VLCS, and~\cite{peng2019moment} for DomainNet.
We report the classification accuracy for each target domain, along with the overall average accuracy for each dataset.

\subsection{Baselines} 
We compare PLAN against 15 recently proposed methods across four categories:

\emph{1) Centralized learning (CL)-based DG methods}
aim to develop a well-generalized model using data from all available domains in a centralized manner.
\begin{itemize}
\item \textbf{SWAD}~\cite{cha2021swad} demonstrates that flatter minima lead to better generalization in unseen domains and proposes a stochastic weight averaging densely algorithm to identify these flatter minima.

\item \textbf{HCVP}~\cite{zhou2024hcvp} improves domain generalization by introducing a hierarchical prompt generation system that captures both domain-specific and task-specific features, which are integrated into a ViT via contrastive learning.

\item \textbf{Doprompt}~\cite{zheng2022prompt} enhances generalization by generating prompts for input images. It first learns domain-specific prompts from source domains, and then trains a prompt adapter to produce tailored prompts for each target image.

\end{itemize}
  
\emph{2) FL-based DG methods} 
enable multiple decentralized clients to collaboratively train a model without directly sharing their data.
\begin{itemize}
\item \textbf{FedAvg}~\cite{mcmahan2017communication} is a classic FL algorithm that derives a global model by averaging locally trained model updates from multiple clients.
\item \textbf{FedProx}~\cite{li2020federated} extends FedAvg by incorporating a proximal term into the objective function to address challenges such as statistical heterogeneity among client data.
\end{itemize}

\emph{3) Conventional FedDG methods} 
combine the principles of FL and DG to develop models that generalize well to unseen domains while ensuring data privacy.
\begin{itemize}
\item \textbf{FedSR}~\cite{nguyen2022fedsr} employs both $L_2$-norm and conditional mutual information regularizations to achieve a simple yet generalizable data representation. 

\item \textbf{FedADG}~\cite{zhang2021federated} adopts adversarial learning to align data distributions across source domains with a dynamically generated reference distribution. 

\item \textbf{CCST}~\cite{chen2023federated} transfers domain-specific styles across clients, allowing the model to capture a broader range of domain variations.

\item \textbf{ELCFS}~\cite{liu2021feddg} facilitates privacy-preserving data exchange among clients by transmitting information in the frequency space, thereby enhancing model generalizability.

\item \textbf{GA}~\cite{zhang2023federated} dynamically recalibrates the aggregation weights of local models by minimizing the variance in generalization gaps across clients.
\end{itemize}

\emph{4) Parameter-efficient fine-tuning (PEFT)-based methods} 
leverage techniques such as prompt learning and adapters to enhance the generalization of VLMs in FL scenarios.
\begin{itemize}
  \item \textbf{FedCLIP}~\cite{lu2023fedclip} incorporates an attention-based adapter to tailor VLMs to individual clients.
  \item \textbf{PromptFL}~\cite{guo2023promptfl} uses local data to train text prompts, which are then averaged to construct a global prompt.
  \item \textbf{FedAPT}~\cite{su2024federated} generates personalized text prompts for test samples by collaboratively training a meta prompt and adaptive network across clients.
  \item \textbf{FedPR}~\cite{feng2023learning} optimizes local visual prompts within the null space of the global prompt to preserve prior global knowledge while updating client-specific parameters.
  \item \textbf{FedMaPLe} replicates MaPLe~\cite{khattak2023maple} in the FL setting, performing both text and visual prompt learning while generating global prompts through averaging. 
\end{itemize}
\subsection{Implementation Details}
In our implementation of PLAN, we utilized the pre-trained ViT-Base/16 CLIP model for prompt learning, with text and visual prompt token dimensions set to 512 and 768, respectively.
The prompt token length was configured to be 8, while the depth of the Transformer blocks incorporating the prompts was fixed at 12.
The mapping functions, $\mathcal{F}_q$ and $\mathcal{F}_a$ in Eq.~\eqref{eq:atten} and Eq.~\eqref{eq:aggre}, were implemented using a two-layer bottleneck network with a reduction ratio of $1/8$.

During training, we set the number of federated rounds to 20 and the number of local update epochs to 1, i.e., $R=20$ and $E=1$ in Algorithm~\ref{alg:training}.
We adopted the SGD optimizer with a batch size of 32 and a learning rate of 0.0015.
The trade-off hyperparameter $\alpha$ in Eq.~\eqref{eq: t_ce} was selected as 1.
The influence of key hyperparameters on model performance will be discussed in a later section.
For the sake of reproducibility, the code of our PLAN method has been made publicly available at~\url{https://github.com/GongShuai8210/PLAN}.

For CL-based DG baselines, we cited the results using the ViT backbone from the previously published paper~\cite{zhou2024hcvp}.
For the remaining three categories of baselines, we reproduced their results based on CLIP to ensure a fair comparison.
The hyperparameters of baseline methods were determined either by following the settings in the original papers or according to the results obtained on the validation set.


\begin{table}[t]
	\begin{center}
		
		\caption{Performance comparison in accuracy (\%) on PACS.}\label{pacs_comparasion}
		\resizebox{0.453\textwidth}{!}{
			\begin{tabular}{c|c|c|c|c|c}
					\toprule
					\multirow{2}{*}{Methods}  & \multicolumn{5}{c}{PACS}\\
					\cmidrule{2-6} 
					& {Art} & {Cartoon} & {Photo} & {Sketch} & {Avg.}\\
					\midrule
					\multicolumn{6}{>{\columncolor{LightBlue}}c}{\textit{CL-based DG methods}}\\
					
					SWAD &93.23 &85.93 &99.18 &82.03&90.44\\
					HCVP &93.17 &86.89 &99.33 &81.30  &90.17 \\ 
					Doprompt &95.00  &86.35  &99.63  &78.20  &89.91 \\
					\midrule
					\multicolumn{6}{>{\columncolor{LightOrange}}c}{\textit{FL-based DG methods}}\\
					FedAvg   &82.67	&65.27	&97.42	&65.18	&77.64 \\
					FedProx   	&83.89	&68.39	&97.43	&64.21	&78.48 \\
					\midrule
					\multicolumn{6}{>{\columncolor{LightGreen}}c}{\textit{Conventional FedDG methods}}\\
					FedSR   &88.19	&67.45	&95.74	&65.92	&79.33 \\
					FedADG   &82.93	&65.42	&98.09	&65.36	&77.95\\
					CCST   	 &87.02	&74.57	&98.29	&65.84	&81.43	\\
					ELCFS   &86.77	&73.21	&98.14	&65.16	&80.82\\
					ELCFS+GA  &87.68	&75.19	&97.56	&65.86	&81.57 \\
					\midrule
					\multicolumn{6}{>{\columncolor{LightRed}}c}{\textit{PEFT-based methods}}\\
					FedCLIP   	&96.19	&97.91	&99.76	&85.85	&94.93\\
					PromptFL   &96.34	&98.46	&99.58&	92.19&	96.64\\
					FedAPT  &97.15	&99.12	&99.69&	\textbf{92.34}&	97.08\\
					FedPR  	&98.10	&99.02	&99.88&	91.11&	97.03\\
					FedMaPLe  &98.44  &99.02 &\textbf{99.94} &90.40&96.95\\
					
					\textbf{PLAN (Ours)} &\textbf{98.58}	&\textbf{99.14}	&99.82	&92.08&	\textbf{97.40}\\
					\bottomrule
				\end{tabular}
			}
		\end{center}
		
	\end{table}		
	
\begin{table}[t]
	\begin{center}
		\caption{Performance comparison in accuracy (\%) on OfficeHome.}\label{officehome_comparasion}
		\resizebox{0.45\textwidth}{!}{
			\begin{tabular}{c|c|c|c|c|c}
					\toprule
					\multirow{2}{*}{Methods}  & \multicolumn{5}{c}{OfficeHome}\\
					\cmidrule{2-6} 
					&{Art} & {Clipart} & {Product} & {Real} & {Avg.}\\
					\midrule
					\multicolumn{6}{>{\columncolor{LightBlue}}c}{\textit{CL-based DG methods}}\\
					SWAD &76.26 &68.87 &86.74 &87.03 &79.73 \\
					HCVP &81.77 &69.76 &88.01 &90.62&82.54 \\ 
					Doprompt  &80.95 &70.88 &88.94 &90.10 &82.72 \\ 
					\midrule
					\multicolumn{6}{>{\columncolor{LightOrange}}c}{\textit{FL-based DG methods}}\\
					FedAvg   &69.29&48.22&	72.54&78.68&67.18 \\
					FedProx   	& 70.71	&48.69&	72.02&	78.51&	67.48 \\
					\midrule
					\multicolumn{6}{>{\columncolor{LightGreen}}c}{\textit{Conventional FedDG methods}}\\
					FedSR   &69.12 &49.69	&72.71	&79.12	&67.66 \\
					FedADG   &69.31&48.76&	72.89&79.13& 67.52\\
					CCST   	&69.23 &51.36 &72.09 &81.27 &68.19\\
					ELCFS   &68.17&50.52&	71.44 &80.11 &67.56\\
					ELCFS+GA  &68.62&50.60&73.35&81.23 &68.45 \\
					\midrule
					\multicolumn{6}{>{\columncolor{LightRed}}c}{\textit{PEFT-based methods}}\\
					FedCLIP   	&78.45	&64.77 &87.68&	87.84&	79.69\\
					PromptFL   &82.98	&68.98 &92.14&	90.27&	83.59\\
					FedAPT  &83.96	&71.65 &91.93 &90.51 &84.51\\
					FedPR  	&84.04	&71.63&81.29&91.34&	84.58\\
					FedMaPLe  &84.56	&72.82&	92.38&	91.07	&85.21\\
					\textbf{PLAN (Ours)} &\textbf{86.65}	&\textbf{74.73}&	\textbf{93.47}	&\textbf{92.06}&\textbf{86.73}\\
					\bottomrule
				\end{tabular}
			}
		\end{center}
	\end{table}

\subsection{Main Results}
Table~\ref{pacs_comparasion}, Table~\ref{officehome_comparasion}, and Table~\ref{vlcs_comparasion} present the results of PLAN compared to the baseline methods across different categories on PACS, OfficeHome, and VLCS, respectively. 
Given the large scale of DomainNet, reproducing all models on this dataset is prohibitively expensive. 
Therefore, only the results of PEFT-based methods on DomainNet are shown in Table~\ref{domainnet_comparasion}.
All results represent the average of three runs, with the best performance in each generalization task highlighted in bold.
	
\subsubsection{Compared with CL-based DG methods}
On different datasets, CL-based DG methods exhibit superior performance, significantly surpassing FL-based DG methods as well as conventional FedDG methods.
However, this improvement comes at the expense of data privacy, as CL-based DG  methods require a centralized framework with unrestricted access to data from all clients.
Our PLAN method addresses this issue by leveraging VLM prompts as a privacy-preserving bridge to facilitate the transfer of domain knowledge among clients.
PLAN considerably outperforms CL-based DG methods with average improvements of 6.96\%, 4.01\%, and 4.21\% on PACS, OfficeHome, and VLCS, respectively.	
\subsubsection{Compared with FL-based DG methods}
For FL-based DG methods, it is evident that FedAvg performs relatively poorly across different datasets, likely due to its equal weighting of all clients’ knowledge, which overlooks the unique characteristics of each domain.
While FedProx enhances FedAvg by addressing the challenge of data heterogeneity, its focus remains on the heterogeneity within the federation of clients, without accounting for generalization to unseen target domains. 
As a result, its performance significantly lags behind that of our PLAN method. 

\subsubsection{Compared with conventional FedDG methods}
Conventional FedDG methods typically offer only marginal performance improvements over standard FL algorithms.
Among them, ELCFS and CCST enable the sharing of sensitive information directly extracted from local data between clients, leading to comparatively strong results.
Further performance gains can be achieved by integrating ELCFS with GA.
Notably, there is a significant performance gap between conventional FedDG methods and our PLAN method.
PLAN outperforms these methods by at least 15.83\%, 18.28\%, and 6.11\% in classification accuracy on PACS, OfficeHome, and VLCS, respectively.
\renewcommand{\arraystretch}{1.03} 
\begin{table}[t]
	\begin{center}
		\caption{Performance comparison in accuracy (\%) on VLCS.}\label{vlcs_comparasion}
		\resizebox{0.45\textwidth}{!}{
			\begin{tabular}{c|c|c| c| c|c}
					\toprule
					\multirow{2}{*}{Methods}  & \multicolumn{5}{c}{VLCS}\\
					\cmidrule{2-6} 
					& {Caltech} & {LabelMe} & {VOC} & {SUN} & {Avg.}\\
					\midrule
					\multicolumn{6}{>{\columncolor{LightBlue}}c}{\textit{CL-based DG methods}}\\
					
					SWAD & 98.49 &68.36 &75.40  &79.49  &79.31  \\
					HCVP &96.32  &66.26  &80.08  &81.65  &81.08   \\
					Doprompt &96.70  &66.53  &78.28  &79.39  &80.23  \\ 
					\midrule
					\multicolumn{6}{>{\columncolor{LightOrange}}c}{\textit{FL-based DG methods}}\\
					FedAvg   &95.48	&65.63&	78.21&	73.31&	78.16\\
					FedProx   &95.41&	64.91&	76.45&	77.9&	78.70
					\\ 
					\midrule
					\multicolumn{6}{>{\columncolor{LightGreen}}c}{\textit{Conventional FedDG methods}}\\
					FedSR   &95.16	&65.86	&78.51	&73.49	&78.26\\ 
					FedADG   &95.21	&65.76	&76.43	&75.96	&78.34\\
					CCST   &96.49&65.73	&76.42	&77.67	&79.08\\
					ELCFS   &95.67	&65.02	&76.55	&77.96	&78.80\\
					ELCFS+GA  &96.77&65.16	&78.89	&78.93	&79.18\\
					\midrule
					\multicolumn{6}{>{\columncolor{LightRed}}c}{\textit{PEFT-based methods}}\\
					FedCLIP   &\textbf{99.93}	&66.98	&73.28	&87.14	&81.83\\
					PromptFL   &99.65	&68.03	&72.24	&85.10&	81.26\\
					FedAPT  &99.71	&68.42	&79.64&	85.56&83.33	\\
					FedPR  &99.36	&68.18	&81.06&	85.98&	83.64\\
					
					FedMaPLe  &98.02	&69.50	&82.15	&85.81	&83.87\\ 
					\textbf{PLAN (Ours)} &99.18	&\textbf{69.94}	&\textbf{83.75}&	\textbf{88.28}	&\textbf{85.29}\\
					\bottomrule
				\end{tabular}
			}
		\end{center}
		
	\end{table}	

\begin{table}[t]
	\begin{center}
		\caption{Performance comparison in accuracy (\%) on DomainNet.}\label{domainnet_comparasion}
		\resizebox{0.5\textwidth}{!}{
			\begin{tabular}{c|c|c|c|c|c|c|c}
					\toprule
					\multirow{2}{*}{Methods}  & \multicolumn{7}{c}{DomainNet}\\
					\cmidrule{2-8} 
					& {Clipart} & {Info} & {Paint} & {Quick} & {Real}&{Sketch}&{Avg.}\\
					\midrule
					\multicolumn{8}{>{\columncolor{LightRed}}c}{\textit{PEFT-based Methods}}\\
					FedCLIP  &74.12	&48.36	&68.49&	31.73&80.52&58.62&60.31\\
					PromptFL   &76.53	&51.72	&70.86	&34.21	&81.68&68.37&63.90\\
					FedAPT  &77.02	&51.45	&70.36&\textbf{49.62}&86.64	&68.43&67.25\\
					FedPR  &75.49	&51.96	&71.42&	35.98&	82.67&69.43&64.49\\
					FedMaPLe  &78.61	&65.23	&71.89	&43.46	&86.32&72.46&69.67\\ 
					\textbf{PLAN (Ours)} &\textbf{79.51}&\textbf{66.42}	&\textbf{72.11}&48.83	&\textbf{86.72}&\textbf{72.69}&\textbf{71.05}\\
					\bottomrule
				\end{tabular}
			}
		\end{center}
	\end{table}
\subsubsection{Compared with PEFT-based methods}
Overall, PEFT-based methods demonstrate significant superiority over other categories of methods across all datasets. 
This is primarily because PEFT-based methods introduce only a small set of additional parameters while keeping the parameters of CLIP unchanged, thereby fully leveraging the original powerful transferability of VLMs.
PLAN is also categorized as a PEFT-based method.
In the following, we provide a detailed analysis of how PLAN compares to other competitors within this category on each dataset:

As shown in Table~\ref{pacs_comparasion}, the adapter-based method FedCLIP lags behind other prompt learning methods on PACS.
Meanwhile, all prompt learning methods achieve high accuracy in various generalization tasks, with no substantial differences in performance observed between them.
Nonetheless, PLAN outperforms the others in two of the four target domains and obtains the highest average performance.

From Table~\ref{officehome_comparasion}, we can see that PLAN demonstrates the best performance across all four domains on OfficeHome. 
Compared to FedMaPLe, the leading PEFT-based baseline, PLAN achieves an average performance improvement of 1.52 percentage points on this dataset.

On VLCS, PLAN falls short of attaining the best performance only in the Caltech domain, as shown in Table~\ref{vlcs_comparasion}.
However, the performance gap is of limited significance, as most methods have already surpassed 99\% accuracy in this domain.
PLAN exhibits clear advantages in the other domains, particularly in the VOC and SUN domains, with performance improvements of at least 1.60\% and 1.14\%, respectively.

Table~\ref{domainnet_comparasion} presents the comparison results on DomainNet. 
As observed, PLAN continues to demonstrate its effectiveness for this large-scale dataset, achieving SOTA results in five out of six domains.
It is worth noting that PLAN is the only method to exceed an accuracy of 70\% among all the evaluated approaches.
In comparison to the runner-up method FedMaPLe, PLAN shows an average improvement of 1.38\% across the six domains.

\begin{table}[t]
	\centering
	\caption{Investigation on the contribution of different components.}
	\label{tab:ablation}
	\resizebox{0.45\textwidth}{!}{
	\begin{tabular}{c|c|c|c|c|c|c}
		\toprule
	 $\mathcal{L}_{kl}^k$ & CLIP's ZSI & $\mathcal{A}_t$ & $\mathcal{A}_v$ & PACS  & OfficeHome & VLCS \\ 
		\midrule
		$\checkmark$ & $\checkmark$ & $\checkmark$ & $\checkmark$ & \textbf{97.40}& \textbf{86.73} & \textbf{85.29}  \\
        $\boldsymbol{\times}$ & $\boldsymbol{\times}$ & $\checkmark$ & $\checkmark$ & 96.55 & 85.81 & 85.05\\
		$\checkmark$ & $\boldsymbol{\times}$ & $\checkmark$ & $\checkmark$ & 97.13 & 86.42  & 85.20  \\
		$\boldsymbol{\times}$ & $\boldsymbol{\times}$ & $\boldsymbol{\times}$ & $\checkmark$ & 96.47 & 85.60 & 83.98   \\
		$\boldsymbol{\times}$ & $\boldsymbol{\times}$ & $\checkmark$ & $\boldsymbol{\times}$ & 96.46 & 85.17 & 83.71   \\
		$\boldsymbol{\times}$ & $\boldsymbol{\times}$ & $\boldsymbol{\times}$ & $\boldsymbol{\times}$  & 96.29 & 84.70  & 83.69  \\
		
		\bottomrule
	\end{tabular}
}
\end{table}

\subsection{Ablation Study}
Table~\ref{tab:ablation} presents the average accuracy of several variants of our PLAN method when different key components are omitted.

As described in Eq.~\eqref{eq:kl_loss}, we align the local prompts of different clients with the global prompts by minimizing the KL divergence between their respective prediction distributions.
In this way, local prompts are indirectly synchronized while eliminating the need for data sharing.
The ablation study reveals that the variant without the KL loss $\mathcal{L}_{kl}^k$ exhibits substantial performance drops of 0.85\%, 0.92\%, and 0.24\% on PACS, OfficeHome, and VLCS, respectively.
These results highlight the necessity of introducing a common reference to 
synchronize all local prompts.
Without proper synchronization, local prompts may become biased towards each client’s private data, thereby compromising the generalization ability of the global model obtained through subsequent aggregation.
Furthermore, since the global prompts are unavailable in the first federated round, we leverage the zero-shot inference (ZSI) capability of CLIP to obtain its prediction distribution for calculating $\mathcal{L}_{kl}^k$.
When this step is omitted, i.e., $\mathcal{L}_{kl}^k$ is excluded in the first round, slight performance degradation is observed on different datasets.



Our PLAN method incorporates two attention-based prompt aggregators, $\mathcal{A}_t$ and $\mathcal{A}_v$, to selectively aggregate local text and visual prompts into their corresponding global prompts, respectively.
Replacing either $\mathcal{A}_t$ or $\mathcal{A}_v$ with simple average aggregation reduces the model’s performance, and the removal of both further results in a significant decline of 1.11\% and 1.36\% on OfficeHome and VLCS, respectively.
This phenomenon suggests that selective prompt aggregation plays a critical role in PLAN, with both text and visual prompt aggregation contributing to performance improvement.



\begin{table}[t]
	\centering
	\captionsetup{position=above}
	\caption{Effect of the loss balancing coefficient $\alpha$.}
	\resizebox{0.4\textwidth}{!}{
	\begin{tabular}{l|c|c|c|c|c}
		\toprule
		$\alpha$          & Art & Clipart & Product & Real & Avg. \\
		\midrule
		0            & 85.74   & 73.72 &92.77 &91.39 &85.81   \\
		0.1          & 86.24   & 73.79 &93.11 &91.60 &86.19   \\
		0.5          & 86.32   & 74.00 &93.12 &91.99 &86.36    \\
		1          &\textbf{86.65}  &\textbf{74.73} &\textbf{93.47} &\textbf{92.06} &\textbf{86.73}     \\
		2			 & 86.03   &74.55  &93.33 &91.60 &86.38       \\            
		10    		 &85.95	   &74.82  &92.59 &91.99 &86.34        \\
		\bottomrule
	\end{tabular}
}
	\label{tab:weights}
\end{table}
\subsection{Hyperparameter Analysis}
We evaluate the influence of different hyperparameters on the performance of our PLAN method on OfficeHome. 
\subsubsection{Effect of Loss Weight}
In Eq.~\eqref{eq: t_ce}, the hyperparameter $\alpha$ controls the weight of the KL loss $\mathcal{L}_{kl}^k$ in optimizing the local prompts.
Table~\ref{tab:weights} presents the results achieved by PLAN with different values of $\alpha$.
Notably, even a small weight, such as $\alpha=0.1$, yields an average performance improvement of 0.38\%.
Such an observation reaffirms the importance of the KL loss in PLAN.
The performance gradually improves as $\alpha$ increases from 0.1 to 1, but begins to decline with any further increase. 
This trend is intuitive, as an excessively high value of $\alpha$ may cause the KL loss to dominate the training process, thereby preventing the local prompts from effectively capturing domain-specific knowledge on each client.
\begin{figure}[t]
	\captionsetup[subfloat]{labelfont={scriptsize},textfont={scriptsize}}
	\centering
	\subfloat[\scriptsize Prompt depth]{\includegraphics[width=0.45\linewidth]{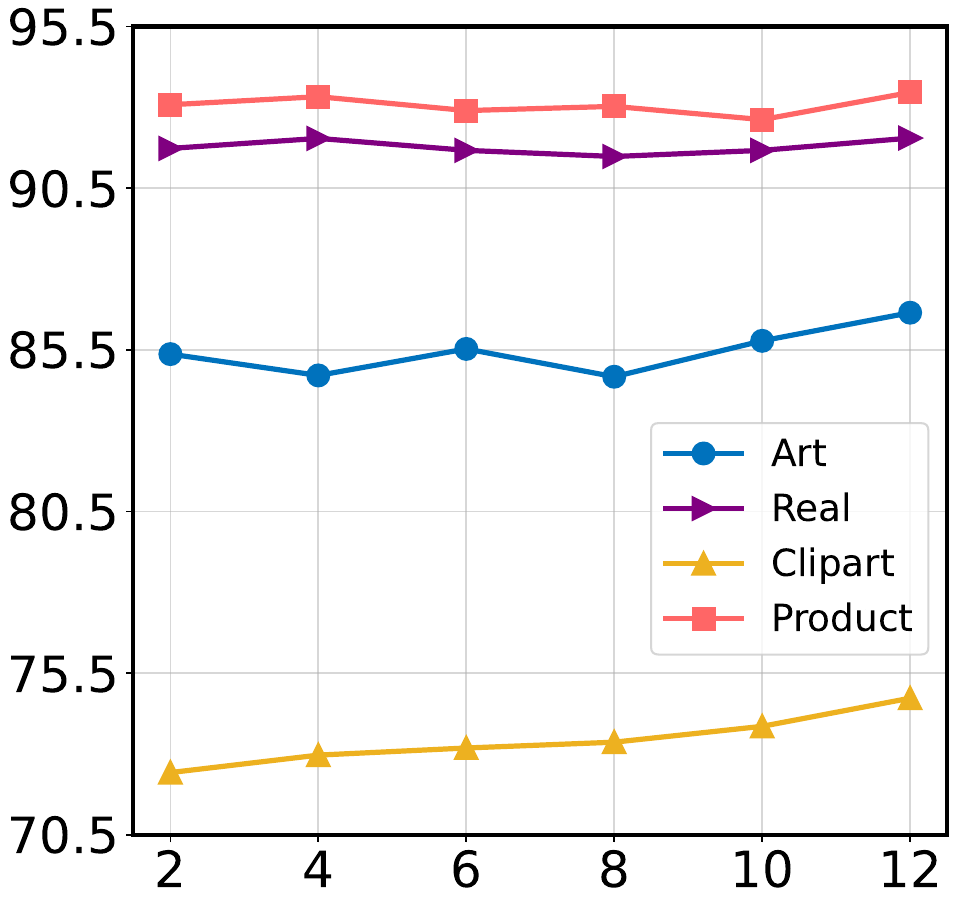}
		\label{p_depth}}
	\subfloat[\scriptsize Prompt length]{\includegraphics[width=0.45\linewidth]{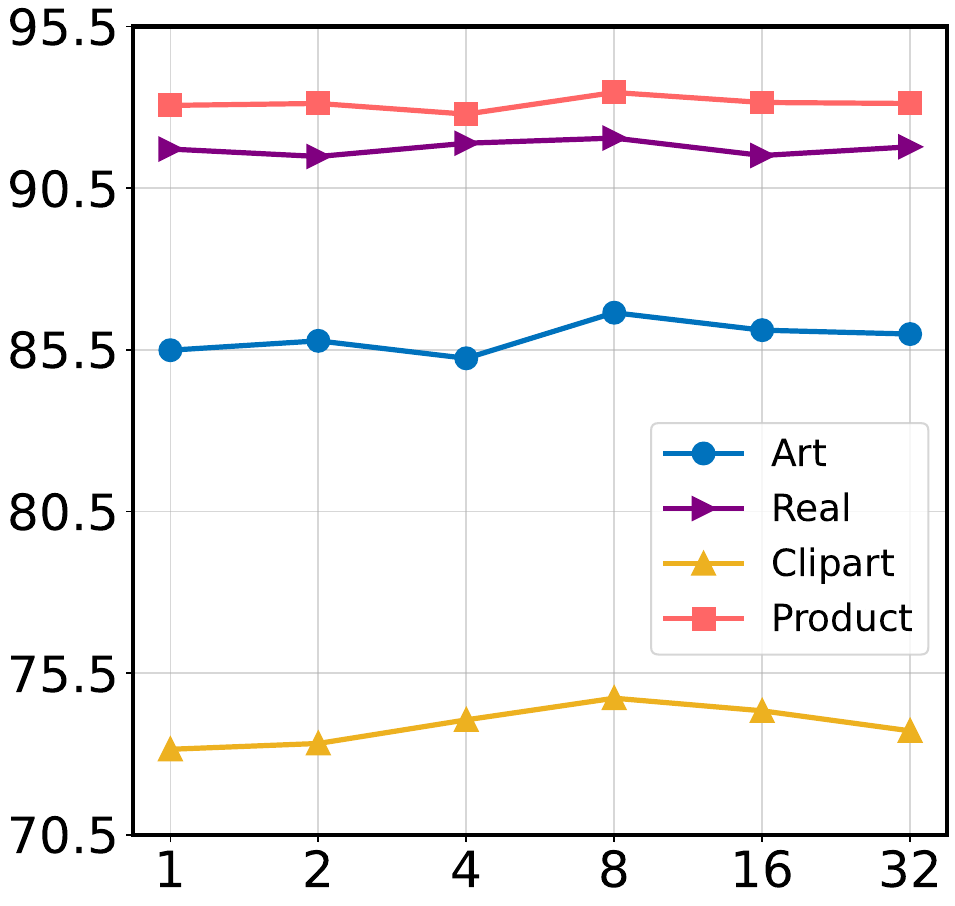}
		\label{p_length}}
	\captionsetup{skip=10pt}
	\caption{Effects of prompt depth and prompt length.}
	\label{fig:prompt}
\end{figure}
\subsubsection{Effect of Prompt Depth and Length}
Fig.~\ref{fig:prompt} displays the effects of adjusting the prompt depth and length on the performance of PLAN. 
For simplicity, we assume that text and visual prompts have identical depth and length~\cite{khattak2023maple}. 
In Fig. \ref{p_depth}, the horizontal axis value $i$ indicates that prompts are inserted into the first $i$ Transformer blocks.
For various target domains, performance generally trends upward as prompt depth increases; however, adding prompts to specific Transformer layers can occasionally result in a slight performance decline.
This phenomenon aligns with the findings from previous studies~\cite{wang2022dualprompt}.
Fig.~\ref{p_length} demonstrates that the optimal results are generally achieved with the prompts containing eight tokens, while further extending prompt length does not necessarily enhance performance.
\begin{figure*}[t]
	\centering
	\includegraphics[width=0.8\linewidth]{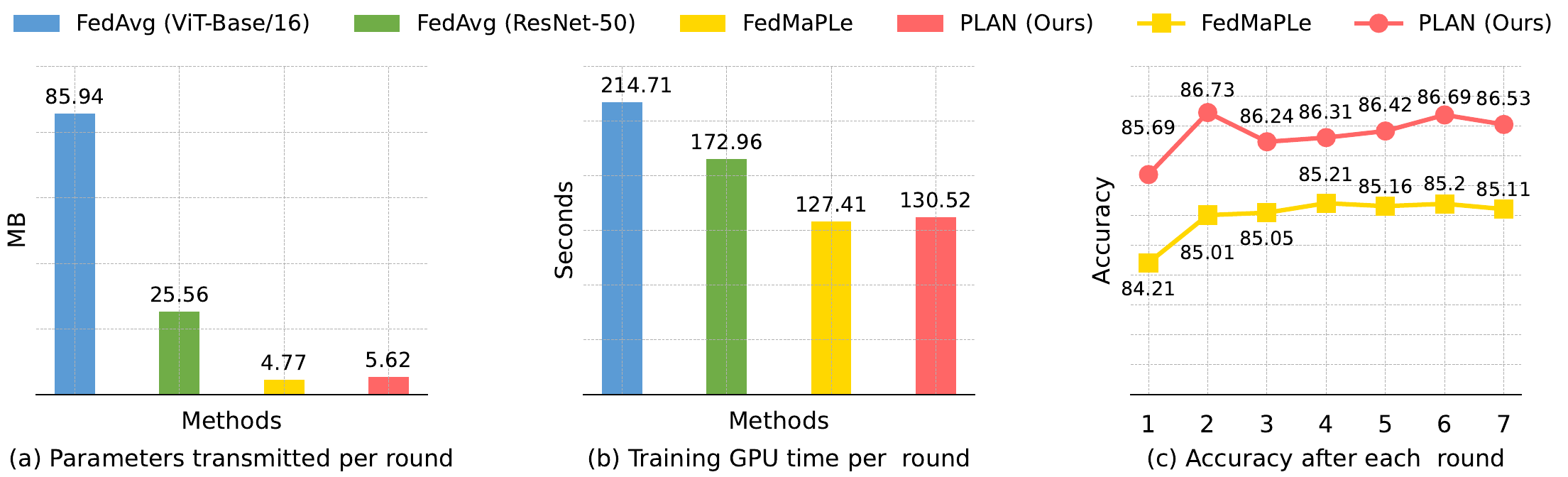}
	\caption{Comparison of computation and communication costs of PLAN and FedDG methods. Each round includes two communications between clients and the server.}
	\label{fig:cost}
\end{figure*}

\vspace{-2pt}
\subsection{Cost Analysis}
\label{cost analysis}
We assess the efficiency of PLAN by examining the communication and computation costs during training on OfficeHome. 
The communication cost is measured by the size of parameters transmitted per federated round, while the computation cost is quantified by the GPU training time required per round.

As shown in Fig.~\ref{fig:cost}\textcolor{red}{a}, we observe that compared to traditional FedDG methods such as FedAvg using ViT-Base/16 and ResNet-50, PLAN reduces the communication cost by up to 15.29 times and 4.55 times, respectively. 
In contrast to the previous leading prompt learning method, FedMaPLe, PLAN incurs a slightly higher communication cost, as it requires two communications between the server and clients in each round. 
Similarly, as illustrated in Fig.~\ref{fig:cost}\textcolor{red}{b}, PLAN exhibits only a slightly lower computational efficiency per round in comparison with FedMaPLe.
However, as depicted in Fig~\ref{fig:cost}\textcolor{red}{c}, when comparing the accuracy of PLAN and FedMaPLe after each round, we find that PLAN consistently achieves considerably higher performance.
Notably, after just one training round, PLAN surpasses FedMaPLe's peak performance over the entire training process.
The faster convergence speed of PLAN significantly reduces its total communication and computational costs relative to FedMaPLe during training. 
\begin{figure}[t]
	\captionsetup[subfloat]{labelfont={scriptsize},textfont={scriptsize}}
	\centering
	\subfloat[\scriptsize PACS]{\includegraphics[width=0.9\linewidth]{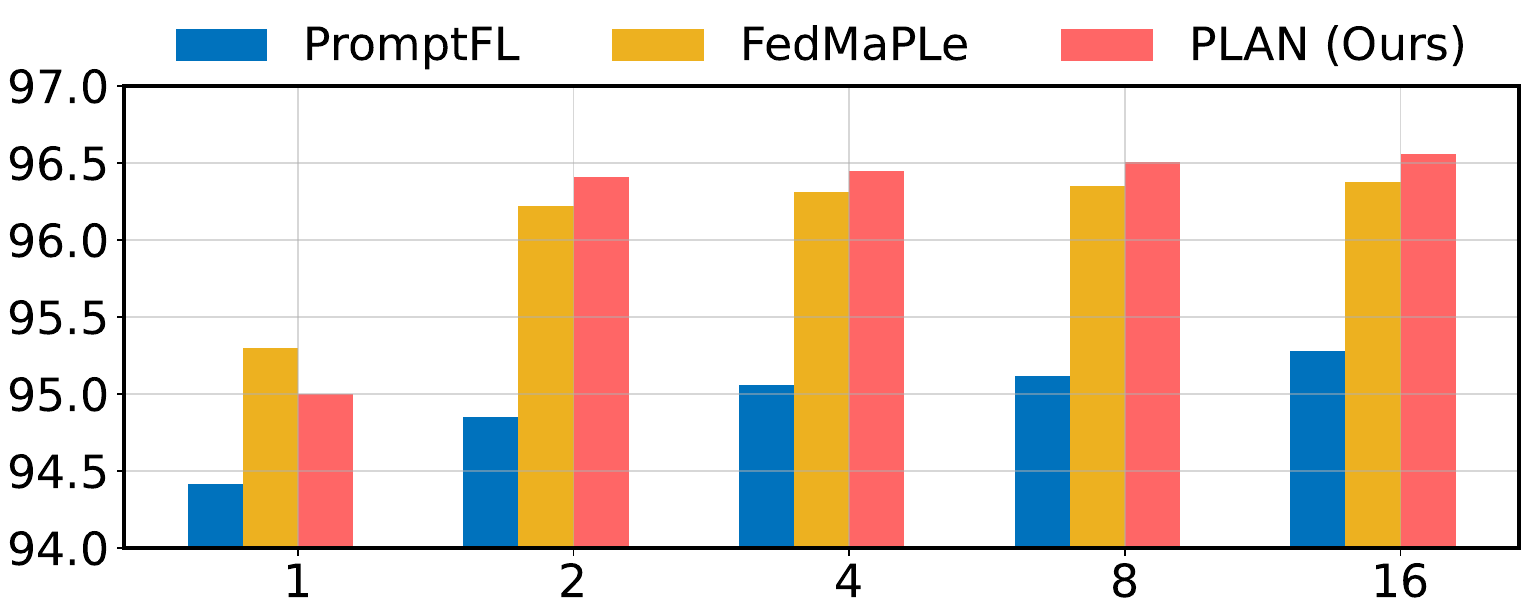}
		\label{pacs_fs}} \\
	\subfloat[\scriptsize OfficeHome]{\includegraphics[width=0.9\linewidth]{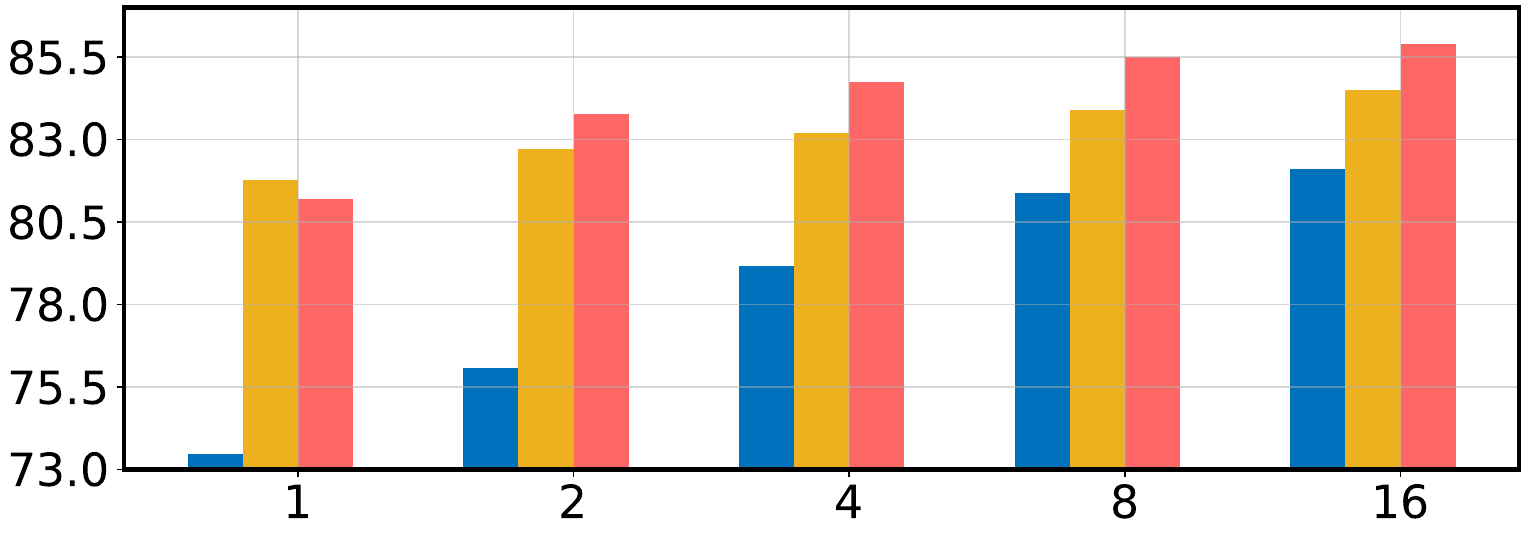}
		\label{officehome_fs}}
	\captionsetup{skip=10pt}
	\caption{Performance comparison in few-shot settings on PACS and OfficeHome. The values on the x-axis represents the number of shots.}
	\label{fig:few_shots}

\end{figure}
\subsection{Comparison in Few-shot Setting}
In FL scenario, each client may have limited data samples. 
To evaluate PLAN's effectiveness in few-shot settings, we compare its average accuracy against that of PromptFL and FedMaPLe on PACS and OfficeHome.
Fig.~\ref{fig:few_shots} shows the results of the three methods across varying numbers of samples per category. 
As expected, the performance of all methods consistently improves with the increase of training data.
Except for the one-shot setting, PLAN demonstrates superior performance compared to both PromptFL and FedMaPLe in all other cases.
While all three methods are based on CLIP, these results demonstrate that PLAN more effectively facilitates the adaptation of CLIP in few-shot settings.

\subsection{Visualization}
We present some visualization results to offer an intuitive understanding of PLAN's advantages.

\begin{figure}[t]
	\captionsetup[subfloat]{labelfont={scriptsize},textfont={scriptsize}}
	\centering
	\subfloat[\scriptsize CLIP]{\includegraphics[width=0.48\linewidth]{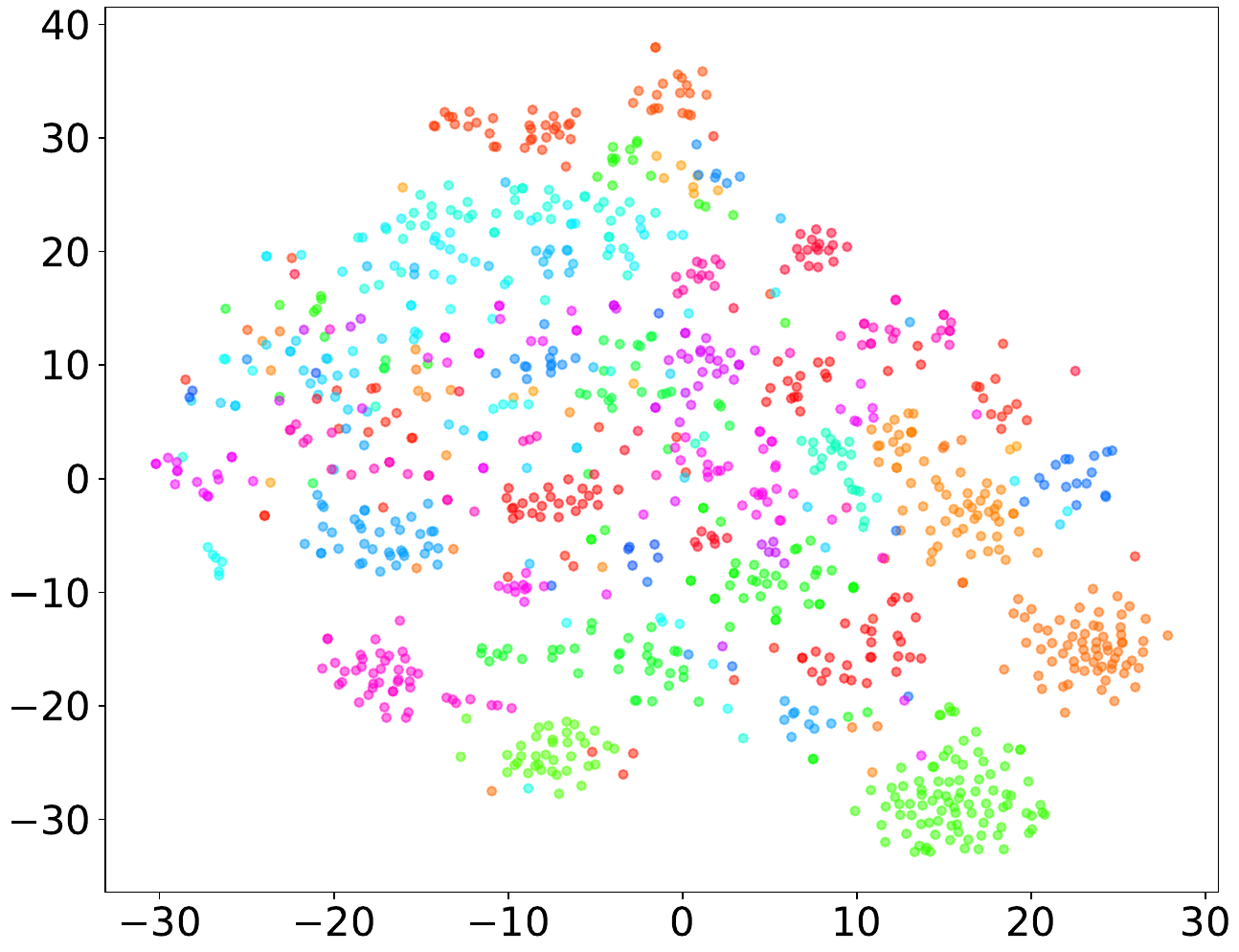}
		\label{f_vlam}}
	\subfloat[\scriptsize PLAN]{\includegraphics[width=0.48\linewidth]{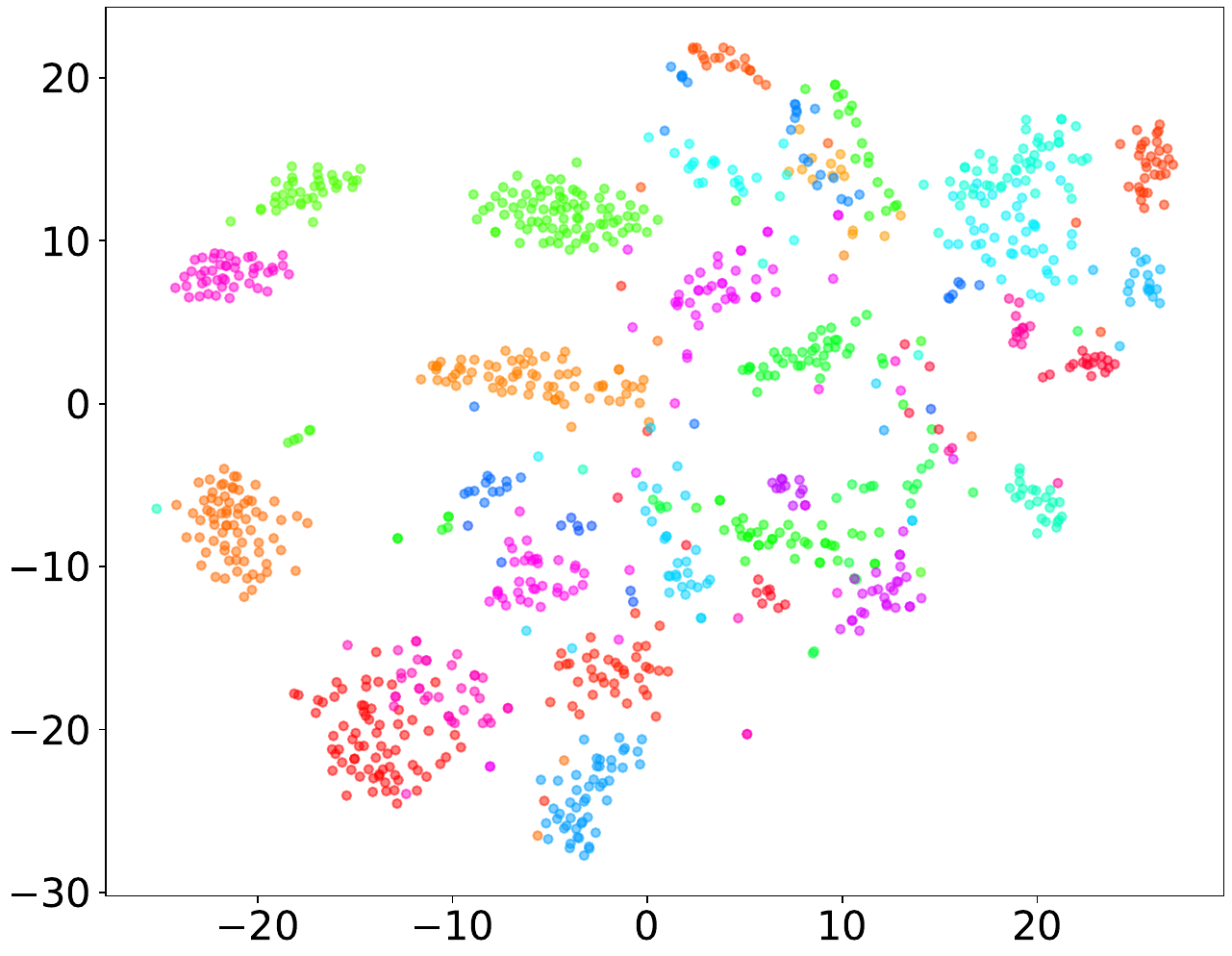}
		\label{f_clip}}
	\captionsetup{skip=10pt}
	\caption{Visualization of the target features extracted by CLIP and PLAN.}
	\label{fig:image_f_vis}
\end{figure}

In Fig.~\ref{fig:image_f_vis}, we apply the t-SNE algorithm~\cite{van2008visualizing} to visualize the features of target samples extracted by CLIP and PLAN, respectively, when ``Real World" is designated as the target domain on OfficeHome.
Samples from the same category are marked with identical colors.
Compared to CLIP, the target features extracted by PLAN are more clustered, and features associated with different categories exhibit greater separation.
This visualization suggests that PLAN achieves a more distinct feature representation across categories.

\begin{figure*}[t]
	\centering
	\includegraphics[width=0.95\linewidth]{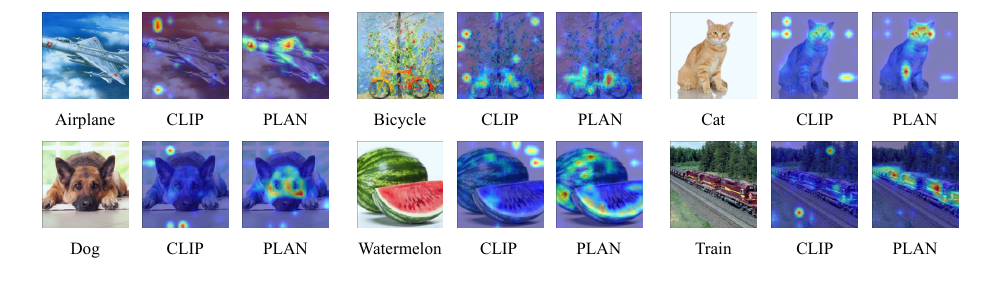}
	\captionsetup{skip=2pt}
	\caption{Visualization of the class activation maps learned by CLIP and PLAN for some target samples.}
	\label{fig:atten}
\end{figure*}
In Fig.~\ref{fig:atten}, we visualize the Class Activation Maps (CAMs)~\cite{zhou2016learning} generated by CLIP and PLAN for some target samples on DomainNet.
To produce the CAMs, we first compute the model's output logits and then use their gradients to weight the attention maps of each block within the vision encoder. 
The final CAMs are obtained by averaging these weighted attention maps across all blocks.
Each sample is annotated with its category label.
It is evident that CLIP frequently focuses on background elements or objects irrelevant to the given category, whereas PLAN effectively concentrates on the objects of interest.

\section{Conclusion}\label{sec:conclusions}

In this paper, we explored the emerging field of federated domain generalization (FedDG) with prompt learning techniques. 
We proposed that learned prompts can act as a more secure bridge for knowledge transfer among clients, as they are not directly generated from local data.
Based on the prompt learning technique in pre-trained vision-language models, we introduced a communication-efficient framework called PLAN, which employs a two-stage collaborative training strategy to generate reference-based local prompts towards each client and  selectively aggregates these local prompts into global prompts. The entire framework optimizes only a small number of parameters for the prompts and lightweight attention-based prompt aggregators, thereby ensuring communication efficiency.
Our extensive experiments on four benchmark datasets demonstrate that PLAN achieves new state-of-the-art performance in FedDG. One limitation of our study is the assumption that the source and unseen domains share the same category space, which might not always hold in real-world applications. Future work will focus on extending PLAN to accommodate open-set scenarios, making it more applicable to practical settings.

\bibliographystyle{IEEEtran}
\bibliography{Ref}

\end{document}